\documentclass[10pt,twocolumn,letterpaper]{article}

\usepackage{cvpr}              % To produce the CAMERA-READY version
\definecolor{cvprblue}{rgb}{0.21,0.49,0.74}
\usepackage[pagebackref,breaklinks,colorlinks,allcolors=cvprblue]{hyperref}

\usepackage{xcolor}  
\usepackage{wrapfig}
\usepackage{duckuments}
\usepackage{CJKutf8}
\usepackage{algorithm}  
\usepackage{algorithmicx}  
\usepackage{algpseudocode}  
\usepackage{overpic}
\usepackage{pifont}
\usepackage{bbding}
\usepackage{comment}
\usepackage{float}

\definecolor{MyDarkBlue}{rgb}{0,0.5,1}
\definecolor{MyDarkGreen}{rgb}{0.02,0.6,0.02}
\definecolor{MyDarkRed}{rgb}{0.8,0.02,0.02}
\definecolor{MyDarkOrange}{rgb}{0.40,0.2,0.02}
\definecolor{MyYellow}{rgb}{1,0.55,0}
\definecolor{MyPurple}{RGB}{111,0,255}
\definecolor{MyRed}{rgb}{1.0,0.0,0.0}
\definecolor{MyGold}{rgb}{0.75,0.6,0.12}
\definecolor{MyDarkgray}{rgb}{0.66, 0.66, 0.66}
\definecolor{default}{RGB}{0,0,0}

\newcommand\tx[1]{\text{#1}}
\newcommand\bb[1]{\textbf{#1}}
\newcommand\ti[1]{\textit{#1}}

\newcommand{\model}{GenReward} % model name
\newcommand{\FB}{forward-backward} % 
\renewcommand{\eqref}[1]{Eq.~(\ref{#1})} %
\newcommand{\figref}[1]{Figure~\ref{#1}} % 
\newcommand{\tabref}[1]{Table~\ref{#1}} %
\newcommand{\appref}[1]{\underline{Supplementary Material~\ref{#1}}} %
\newcommand{\mycheckmark}{{\textcolor{MyDarkGreen}{\checkmark}}}
\newcommand{\myxmark}{{\textcolor{MyDarkRed}{\ding{55}}}}

\def\paperID{*****} % *** Enter the Paper ID here
\def\confName{CVPR}
\def\confYear{2026}

\title{Goal-Driven Reward by Video Diffusion Models for Reinforcement Learning}
\author{
Qi Wang$^{1,2,3}$\footnotemark[1]\thanks{Equal contribution.} \
Mian Wu$^{1}$\footnotemark[1] \
Yuyang Zhang$^{1,2,3}$\footnotemark[1] \
Mingqi Yuan$^{2,3,5}$ \
Wenyao Zhang$^{1,2,3}$ \
Haoxiang You$^{6}$ \\
Yunbo Wang$^{1}$ \
Xin Jin$^{2,3,4}$\thanks{Corresponding author: Xin~Jin~\textless jinxin@eitech.edu.cn\textgreater.} \
Xiaokang Yang$^{1}$\
Wenjun Zeng$^{2,3}$\\
$^1$ Shanghai Jiao Tong University \
$^2$ Ningbo Key Laboratory of Spatial Intelligence and\\Digital Derivative, Ningbo Institute of Digital Twin, Eastern Institute of Technology, Ningbo\\
$^3$ Zhejiang Key Laboratory of Industrial Intelligence and Digital Twin\\
$^4$ Zhongguancun Academy  
$^5$ Department of Computing, The Hong Kong Polytechnic University\\
$^6$ Department of Mechanical Engineering, Yale University\\
\textcolor{magenta}{\url{https://qiwang067.github.io/genreward}}
}

\begin{document}
\maketitle  

\begin{abstract}
Reinforcement Learning (RL) has achieved remarkable success in various domains, yet it often relies on carefully designed programmatic reward functions to guide agent behavior. 
Designing such reward functions can be challenging and may not generalize well across different tasks. 
To address this limitation, we leverage the rich world knowledge contained in pretrained video diffusion models to provide goal-driven reward signals for RL agents without ad-hoc design of reward.
Our key idea is to exploit off-the-shelf video diffusion models pretrained on large-scale video datasets as informative reward functions in terms of video-level and frame-level goals. 
For video-level rewards, we first finetune a pretrained video diffusion model on domain-specific datasets and then employ its video encoder to evaluate the alignment between the latent representations of agent's trajectories and the generated goal videos. 
To enable more fine-grained goal-achievement, we derive a frame-level goal by identifying the most relevant frame from the generated video using CLIP, which serves as the goal state. 
We then employ a learned forward-backward representation that represents the probability of visiting the goal state from a given state-action pair as frame-level reward, promoting more coherent and goal-driven trajectories.
Experiments on  Meta-World and Distracting Control Suite demonstrate the effectiveness of our approach.
\end{abstract}

\section{Introduction}

Reward feedback is crucial for reinforcement learning (RL) agents to learn effective policies. However, designing appropriate reward functions can be challenging and often requires domain expertise and human labor. 
This limitation hinders the scalability and applicability of RL in complex scenarios.
In contrast, humans can learn desired behaviors from demonstrations or high-level instructions without customized reward design.

Previous research has explored various approaches to address this challenge. One popular solution to this challenge is to exploit expert videos or demonstrations to design reward signals.
RoboCLIP~\cite{sontakke2023roboclip} leverages pretrained vision-language models (VLMs) to calculate the similarity between task descriptions or demonstration videos and historical frames as rewards for RL agents.
Diffusion reward~\cite{huang2024diffusion} employs a conditional video diffusion model that pretrained on expert videos and uses the entropy of the predicted distribution as rewards.
TADPoLe~\cite{luo2024text} adopts a pretrained, frozen text-conditioned diffusion model to compute zero-shot rewards for text-aligned policy learning.
However, existing approaches do not exploit the generated videos as goal-driven rewards to transfer the rich world knowledge learned by generative models, limiting their ability to provide effective reward signals in complex tasks.
% for reward computation.
%
%
\begin{figure}[t]
    \centering
    \includegraphics[width=\linewidth]{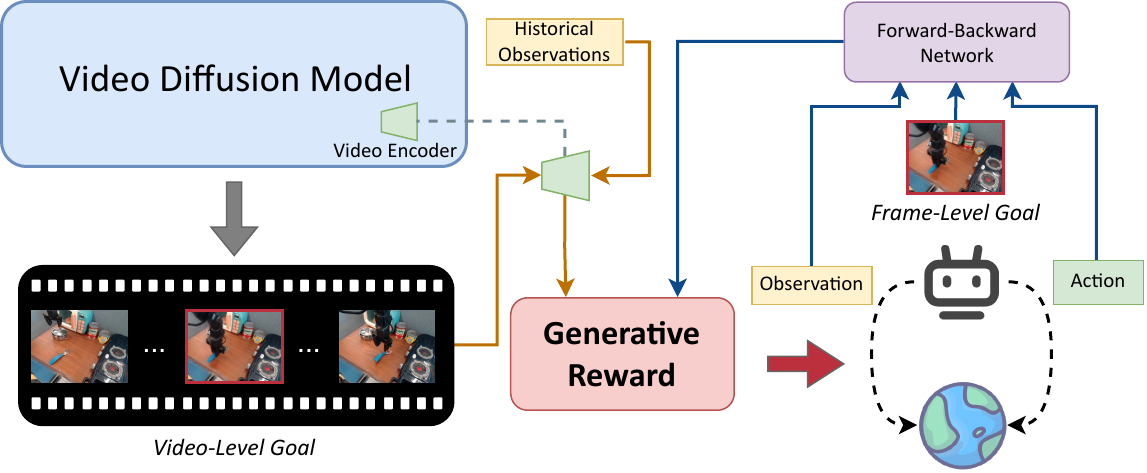}
    \caption{Overview of our proposed framework. The key idea is to leverage generated goal-conditioned videos for world knowledge transfer, enabling the downstream agent to improve performance on unseen tasks.
     }
    \label{fig:overview}
    \vspace{-15pt}
\end{figure}

\begin{table*}[t!]
\caption{Compared to other competitive reward models, the proposed reward framework is based on generative models, does not require expert demonstrations, and incorporates action information for fine-grained goal-achievement.} 
\label{tab:setting_cmp}
\vspace{-10pt}
\setlength\tabcolsep{5pt}
\begin{center}
% \begin{small}
\small
% \scriptsize
\centering
\begin{tabular}{lccc}
\toprule
Model & Demo Free? & Generated?&  Action-aware? \\
\midrule
RoboCLIP~\cite{sontakke2023roboclip} & \myxmark &\myxmark & \myxmark \\
VLM-RMs \cite{rocamonde2024vision} & \mycheckmark &\myxmark & \myxmark \\
LIV~\cite{ma2023liv} & \mycheckmark &\myxmark & \myxmark \\
VIPER~\cite{escontrela2023video} & \myxmark   &\mycheckmark  & \myxmark \\
Diffusion Reward~\cite{huang2024diffusion} & \myxmark   &\mycheckmark  & \myxmark \\
TADPoLe~\cite{luo2024text} & \mycheckmark &  \mycheckmark  & \myxmark \\
\model{}~(Ours) & \mycheckmark &  \mycheckmark  & \mycheckmark \\
\bottomrule
\end{tabular}
\end{center}
\vspace{-10pt}
\end{table*}

In this paper, we propose a novel reward framework that leverages pretrained video diffusion models as goal-driven reward models for RL agents, dubbed Generative Reward~(\model{}).
Concretely, as illustrated in \figref{fig:overview}, we first utilize a finetuned video diffusion model to generate goal-conditioned videos based on task descriptions. 
Subsequently, to achieve the video-level goal, we employ the video encoder of the pretrained generative model to extract latent representations from both the agent's observations and the generated goal videos. We then calculate the correlation between these two latents as a video-level reward.
Meanwhile, for fine-grained goal-reaching, we learn \ti{forward-backward} (FB) representation that measures the probability of reaching the goal state that is selected using CLIP from a given state–action pair, which serves as a frame-level reward to achieve the frame-level goal.

\begin{figure*}[t]
    \centering
    \includegraphics[width=\linewidth]{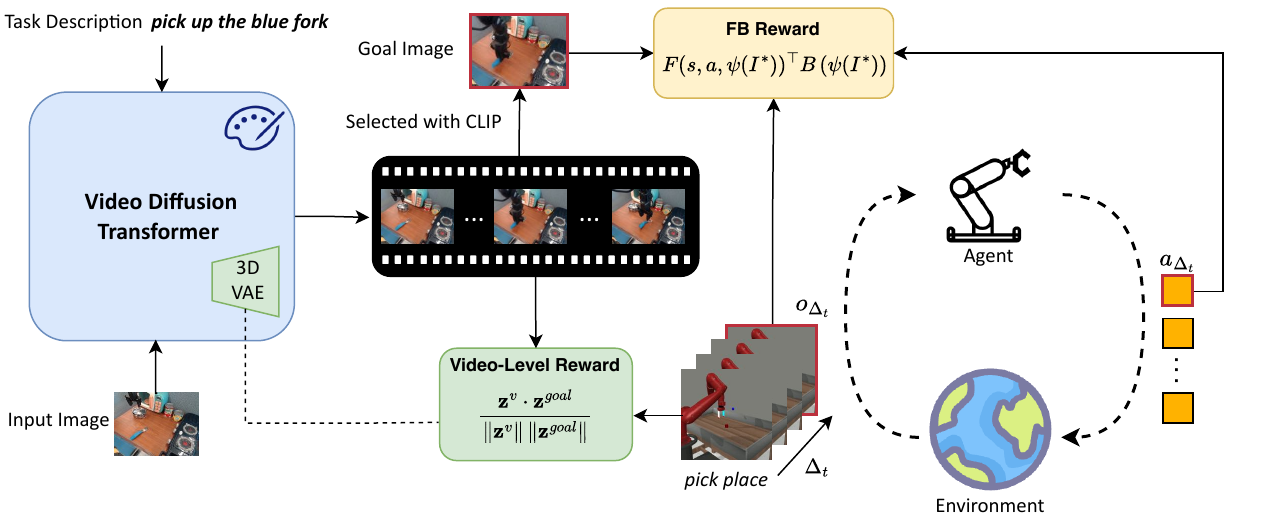}
    \vspace{-15pt}
    \caption{    
    Pipeline of \model{}, which computes goal-driven rewards for behavior learning of the agent using generative prior. 
    During online interaction with the environment, at regular intervals, we employ the correlation between the latent representations of the agent's observations and the generated goal videos as video-level rewards.
    Meanwhile, we learn a forward-backward model to measure the probability of reaching the goal state that is selected using CLIP from a given state–action pair, providing frame-level reward for fine-grained goal-achievement.
% The frame-level goal is selected from the generated video using CLIP, and DINOV3 is adopted to obtain the goal state representation.
    }
    \label{fig:pipeline}
\end{figure*}

Experiments on Meta-World and Distracting Control Suite demonstrate the effectiveness of our approach. Our results show that \model{} significantly outperforms existing methods in terms of episode return.
The contribution of our work can be summarized as follows:
\begin{itemize}
    \item We propose a novel reward framework that exploits pretrained video diffusion models as goal-driven rewards in terms of video-level and frame-level goals, which enables RL agents to receive informative reward signals without handcrafted design.
    \item We incorporate the action information to introduce \FB{} representation as frame-level rewards, which encourages action that is more likely to reach the goal state from a given state–action pair for fine-grained goal-achievement.
\end{itemize}

% \vspace{-3pt}
\section{Problem Setup}
\label{sec:problem_setup}
% \vspace{-3pt}
We solve online visual reinforcement learning as a partially observable Markov decision process (POMDP).
At each timestep $t$, the agent receives an observation $o_t \in \mathcal{O}$ from the environment, takes an action $a_t \in \mathcal{A}$ according to its policy $\pi(a_t|o_t)$, and then transitions to the next observation $o_{t+1}$ while receiving a reward feedback $r_t$.
Concretely, we focus on the scenario where a pretrained video diffusion model that contains rich prior world knowledge is accessible. 
The goal is to improve the online performance of the agent on downstream tasks by leveraging the generative prior to provide intrinsic rewards $r_t^\tx{intr}$.
In comparison, as shown in \tabref{tab:setting_cmp}, existing reward models present notable distinctions in learning paradigms, \textit{i.e.}, the use of expert demonstrations, the reliance on a generative model, and the incorporation of action information.

% \vspace{-3pt}
\section{Method}

\begin{algorithm*}[t]      
    \caption{The training pipeline of \model{}.}
    \label{algo:overall}
    \begin{algorithmic}[1]
    \small
    \State \textbf{Initialize:} World model $\mathcal{M}_\phi$, policy $\pi_\psi$, value function $v_\xi$.
    \State \textbf{Load} pre-trained video diffusion model $G_\theta$.
    \For{finetuning step $t=1, 2, \ldots, K_1$} \Comment{Finetune Video Diffusion Model}
        \State Sample batch $\mathcal{D}_{\text{batch}} \sim \mathcal{D}_{\text{video}}$. 
        \State Update $\theta$ by minimizing the video diffusion model loss on $\mathcal{D}_{\text{batch}}$ using \eqref{eq:vdt_loss}.
    \EndFor
    \State Train the random agent and collect a replay buffer $\mathcal{B}$.
    \While{not converged}
        \Comment{A. Model and Behavior Learning}
        \State Update world model and learn behavior in imaginary trajectories.
        \State $o_1 \leftarrow \text{env.reset()}$. 
        \State Select target goal image from video $\mathbf{V}^{\text{goal}}$ generated with the prompt using CLIP. 
        \State Train \FB{} representation with transitions in $\mathcal{B}$ using \eqref{eq:fb_loss}.
        \For{timestep $t=1, 2, \ldots, T$} \Comment{B. Interacting with the Environment and Reward Shaping}
            \State $a_t \sim \pi_{\psi}(a_t \mid o_t)$.
            \State $o_{t+1}, r_t^{\text{env}} \leftarrow \text{env.step}(a_t)$.
            \If{$t \bmod \tx{intetval } \Delta_t = 0$ }
            \State Compute video-level and \FB{} rewards using \eqref{eq:video_level_reward} and \eqref{eq:fb_reward}, respectively.
            \State Compute generative reward $r_t^{\text{gen}}$ using \eqref{eq:gen_reward}.
            \State $\mathcal{B} \leftarrow \mathcal{B} \cup \{(o_t, a_t, r_t^{\text{gen}}, o_{t+1})\}$.
            \Else
            \State $\mathcal{B} \leftarrow \mathcal{B} \cup \{(o_t, a_t, r_t^{\text{env}}, o_{t+1})\}$.
            \EndIf            
        \EndFor
    \EndWhile
  \end{algorithmic}
\end{algorithm*}

In this section, we present the details of \model{} framework, which involves three main stages~(see \figref{fig:pipeline}):
\begin{enumerate}
    \renewcommand{\labelenumi}{\alph{enumi})}
        \item \textit{Video diffusion model adaptation}: Finetune a pretrained video diffusion model~(VDM) with manipulation videos to enable goal-conditioned video generation.
        \item \textit{Video-Level Goal as Reward}: 
        Employ the correlation between the latents of generated goal-conditioned video and historical visual observations using the video encoder of finetuned video diffusion model as rewards to achieve video-level goal.
        \item \textit{Frame-Level Goal as Reward}: 
        Select the most relevant frame from the generated video as the goal image and learn \FB{} representation to encourage the agent to take actions that are more likely to
        achieve frame-level goal.
\end{enumerate}

\subsection{Preliminary: VDM Adaptation} 
To generate goal-conditioned videos for achieving video-level and frame-level goals, we employ CogVideoX~\cite{yang2025cogvideox}, an image-to-video generation model pretrained on large-scale video datasets.
CogVideoX is built on Video Diffusion Transformers (DiTs),  which use a 3D VAE to map video data $\mathbf{V} \in \mathbb{R}^{F \times 3 \times H \times W}$ into a patchified video latent $x$ and train the diffusion within this latent space.
We then finetune the pretrained DiT with task-specific condition $C = \{c_\text{text}, c_\text{image}\}$, where $c_\text{text}$ and $c_\text{image}$ are text and image embedding extracted from prompt.

The diffusion process consists of a forward Markov chain $\{ \mathbf{x}_t \}_{t=1}^T$ that gradually perturbs the latent variable $\mathbf{x}_0$ with Gaussian noise $\boldsymbol{\epsilon}$, defined as

\begin{equation}
\mathbf{x}_t = \alpha_t \mathbf{x}_0 + \sigma_t \boldsymbol{\epsilon}, \quad \boldsymbol{\epsilon} \sim \mathcal{N}(0, \mathbf{I}).
\end{equation}
Subsequently, a corresponding reverse process parameterized by $p_\theta$ that learns to remove the noise step by step. The denoising model $\hat{\boldsymbol{\epsilon}}_\theta$ is optimized using the standard diffusion objective:

\begin{equation}
\min_\theta \, \mathbb{E}_{\boldsymbol{\epsilon} \sim \mathcal{N}(0, \mathbf{I})}
\left\| \hat{\boldsymbol{\epsilon}}_\theta(\mathbf{x}_t, t, c_\tx{text}, c_\tx{image}) - \boldsymbol{\epsilon} \right\|_2^2,
\label{eq:vdt_loss}
\end{equation}
where $t$ is a timestep uniformly distributed between $[1, T]$.
Benefiting from adaptation of the video diffusion model, we can generate task-specific videos that align with the desired goals.

\subsection{Video-Level Goal as Reward} 
To mimic the desired behavior at the trajectory level, we leverage video-level goals as rewards.
Concretely, given a task specification and task-relevant image, we generate a goal video $\mathbf{V}^{\text{goal}}$ that illustrates the expert-level behavior. 
%
% We select a  as the conditioning image.
%
Since the video encoder of the pretrained video diffusion model contains world knowledge can be helpful for video understanding, we adopt a 3D causal VAE that compresses videos into latent representations both spatially and temporally to measure the similarity between the agent's observations and the generated goal video.
The encoder of finetuned 3D Causal VAE in CogVideoX compresses pixel-level inputs into latent representations, and employs Kullback-Leibler regularizer to help the model learn a meaningful and well-structured latent space.
To handle the length mismatch between the goal video and the agent's historical frames, we uniformly sample 16 frames from each sequence.
The sequence of visual observations \( \mathbf{o}_{0:T} \) is encoded into a latent vector \( \mathbf{z}^v \) using the 3D Causal VAE as follows:
\begin{equation}
\mathbf{z}^v=\text{3D Causal VAE}\left(\mathbf{o}_{0: T}\right).
\end{equation}

Similarly, the goal video \( \mathbf{V}_{\text{goal}} \) is also encoded into a latent vector \( \mathbf{z}^{\text{goal}} \):
\begin{equation}
    \mathbf{z}^{\text{goal}}=\text{3D Causal VAE}\left(\mathbf{o}_{0: K}\right).
\end{equation}

We then compute the cosine similarity between the encoded latent vectors of the agent's observations and the goal video as the video-level reward:
\begin{equation}
r^{\text{video}}=\cos \left(\mathbf{z}^v, \mathbf{z}^{\text{goal}}\right)=\frac{\mathbf{z}^v \cdot \mathbf{z}^{\text{goal}}}{\left\|\mathbf{z}^v\right\|\left\|\mathbf{z}^{\text{goal}}\right\|}.
\label{eq:video_level_reward}
\end{equation}

The video-level reward incentivizes the agent to mimic its behavior with the generated goal video temporally, thus facilitating the achievement of video-level goals.
Notably, we use expert video datasets as finetuning datasets. However, our approach can adopt text-to-video generation without any finetuning datasets. Please refer to \appref{sec:eval_benchmarks}.

\subsection{Frame-Level Goal as Reward}

While video-level goals provide a representation of the desired behavior at the trajectory level, they alone are insufficient for training a well-behaved policy.
To capture fine-grained behaviors at the frame level, we incorporate frame-level goals and action information into the reward function.
This is achieved by first extracting the most relevant frame from the generated video and then encouraging the state distribution visited by the control policy to align with the goal state.
For key frame selection, we adopt OpenCLIP~\cite{radford2021learning} to calculate the similarity between the goal video frames and text description, and select the highest-scoring frames as key frames:
\begin{equation}
I^*=\underset{i}{\arg \max } \frac{\text{CLIP}_{L}(G) \cdot \text{CLIP}_{I}(I_i)}{\left\|\text{CLIP}_{L}(G)\right\|\left\|\text{CLIP}_{I}(I_{i})\right\|}.
\end{equation}
Here, $G$ is the task description, $\text{CLIP}_{L}$ and $\text{CLIP}_{I}$ denote the text and image encoder of CLIP, respectively.

For better generalization of frame-level rewards across diverse goal conditions, we learn a forward–backward representation~\cite{touati2023does} that effectively decomposes the long-term state occupancy $M(s, a, s', \pi_z) $~\footnote{Long-term state occupancy represents the probability that a target state $s'$ is visited starting from a given state–action pair $(s,a)$.} under arbitrary policies.
Concretely, we learn two representations, 
$B: S \rightarrow Z$ and $F: S \times A \times Z \rightarrow Z$.
Here, $Z\in\mathbb{R}^d$ is a $d$-dimensional representation space and $z$ is the vector of representation space $Z$.
The corresponding long-term transition probability can be approximated as
\begin{equation}
\label{eq:low_rank_app}
    M(s, a, s', \pi_z) \approx F^\top(s, a, z) B(s') \rho(\mathrm{d}s'),
\end{equation}
for any policy $\pi_z = \underset{a}{\text{argmax}} \ F(s, a, z)^\top z$.
Here, $\rho$ is the distribution of state-action pairs visited in a training dataset.
Intuitively, $F(s, a, z)^\top B(s')$ approximates the long-term probability of reaching state $s'$ from $(s,a)$ if following policy $\pi_z$.

Before training \FB{} representation, we use DINOv3~\cite{simeoni2025dinov3} to encode the goal image into a semantic representation $\psi(I^*)$, while the model state $s\doteq\phi(o)$ is extracted from the observations using a DreamerV3~\cite{hafner2023dreamerv3} encoder.
The extracted feature vectors are then fed into the forward–backward representation module for learning.

We optimize forward-backward representation to enable accurate approximation of long-term state occupancy as detailed in \eqref{eq:low_rank_app}, which minimizes Bellman residual $\mathcal{L}(F,B)$ as follows: 
\begin{equation}
\label{eq:fb_loss}
\begin{aligned}
 & \|F_z^\top B \rho - (P + \gamma P_{\pi_z} \bar{F}_z^\top \bar{B} \rho)\|\\
&=\mathbb{E}_{(s_t,a_t,s_{t+1})\sim \rho}
\Big[
   \big(
   F(s_t,a_t,z)^\top B(\psi(I^*))
\\
&-\gamma\,\bar F(s_{t+1},\pi_z(s_{t+1}),z)^\top
            \bar B(\psi(I^*))
   \big)^2
\Big]
\\[4pt]
&-2\,\mathbb{E}_{(s_t,a_t,s_{t+1})\sim \rho}
   \Big[
   F(s_t,a_t,z)^\top B(\psi(I^*))
   \Big]
\\[4pt]
&+\text{Const},
\end{aligned}
\end{equation}

where $\bar{F}$ and $\bar{B}$ are target networks updated via a slow-moving average to stabilize training, $z$ is sampled from \ti{Gaussian} distribution for exploration.
The constant term is independent of all learnable parameters. 
More details of forward-backward representation can be found in \appref{sec:fb_details}.

\begin{figure}[t]
    \centering
    \includegraphics[trim=0 70 0 0, clip, width=\linewidth]{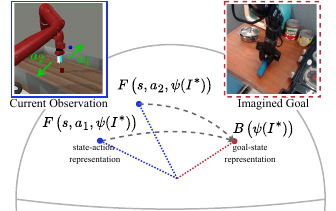}
    \caption{
        Goal-driven action selection. Learned representation space enables goal-directed control by selecting the action whose forward representation of the current state–action pair most closely aligns with the backward representation of goal state.}
    \label{fig:fb_example}
\end{figure}

Once we learned \FB{} representations, we can define frame-level \FB{} reward as follows:
\begin{equation}
r^{\text{FB}}\left(s, a, I^{*}\right)=F(s, a, \psi(I^{*}))^\top B\left(\psi(I^{*})\right).
\label{eq:fb_reward}
\end{equation}
Here, we use $z = \psi(I^*)$ as both the goal embedding for the forward network and the target for computing the reward, keeping consistency with our training objective. 
This reward quantifies how likely the agent can reach the goal state $I^*$ from the current state-action pair $(s, a)$.
As visualized in \figref{fig:fb_example},  we encourage the agent to take actions that are more likely to reach the goal representation.

Finally, we combine our video-level reward and FB reward, and the raw task reward provided by the environment:
\begin{equation}
r^{\mathrm{gen}}=\alpha \cdot r^{\mathrm{video}}+\beta \cdot r^{\mathrm{FB}} + r^{\tx{env}},
\label{eq:gen_reward}
\end{equation}
where $\alpha$ and $\beta$ are weighting coefficients, and $r^{\tx{env}}$ is the environment reward. 

% \vspace{-3pt}
\begin{figure}[t]
    \centering
    \includegraphics[width=\linewidth]{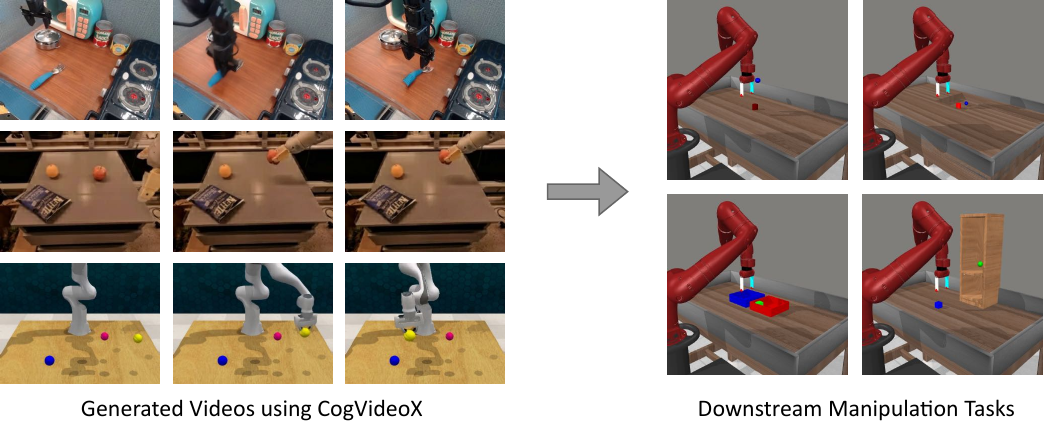}
    \caption{Illustration of experimental setups in our experiments with generated videos and image observations from environments. 
    }
    \label{fig:example_observations}
\end{figure}
\section{Experiments}

\subsection{Experimental Setups} 
\paragraph{Benchmark.} 
We evaluate \model{} on Meta-World~\cite{yu2019meta} and Distracting Control Suite~(DCS)~\cite{stone2021distracting}. Meta-World is a widely adopted benchmark with comprehensive and flexible robotic-control tasks. 
Following \cite{seo2023masked}, we evaluate five medium- or hard-level manipulation tasks in the Meta-World benchmark, including \textit{Pick Place}, \textit{Pick Out of Hole}, \textit{Bin Picking}, \textit{Shelf Place}, and \ti{Disassemble}.
To evaluate the ability of \model{} to learn useful world knowledge from generated videos, we use videos collected in robotic manipulation tasks from RT-1~\cite{brohan2022rt} and Bridge dataset~\cite{walke2023bridgedata}, and Robot Learning Benchmark~(RLBench)~\cite{james2020rlbench} dataset curated by~\cite{zhen2025tesseract} as source domains of generated goal videos (see \figref{fig:example_observations}). 
To further increase the difficulty of the tasks, the length of the episode is limited to 256 steps.
DCS is a modified DeepMind Control Suite that~\cite{tassa2018deepmind} introduces different types of visual distractors. Specifically, we use variations in the background.
The corresponding results on DCS are provided in \appref{sec:eval_benchmarks}.
Notably, we evaluate the models with the original dense reward and sparse reward provided by the environment.

\paragraph{Implementation details.}
We utilize the pretrained CogVideoX-5B-I2V video generation model as a visual prior for goal-conditioned learning. The 3D VAE encoder of CogVideoX-5B-I2V maps RGB videos into a compressed latent space.
Concretely, the demonstration video generated by CogVideoX-5B-I2V at 480$\times$480 resolution, then encoded it into a 16-frame latent sequence using the VAE encoder. 
The shape of each latent frame is (4, $h$/8, $w$/8), where $h$ and $w$ denote the original spatial dimensions. 
During online interaction, every 128 steps, historic frames are encoded with the 3D VAE encoder to obtain the current latent sequence. 
The latent representation is flattened, and the cosine similarity between the current latent and the goal latent is employed as the video-level reward. 
The goal image is encoded into a 384-dimensional goal feature using a frozen DINOv3 model.
During the initial 100k steps, the FB network is trained using transitions sampled from the replay buffer. 
Subsequently, the parameters of the FB network are frozen and the network is used to compute frame-level rewards, thereby stabilizing reward estimation in the later stages of training.

\begin{figure*}[t]
    \centering
    \includegraphics[width=\linewidth]{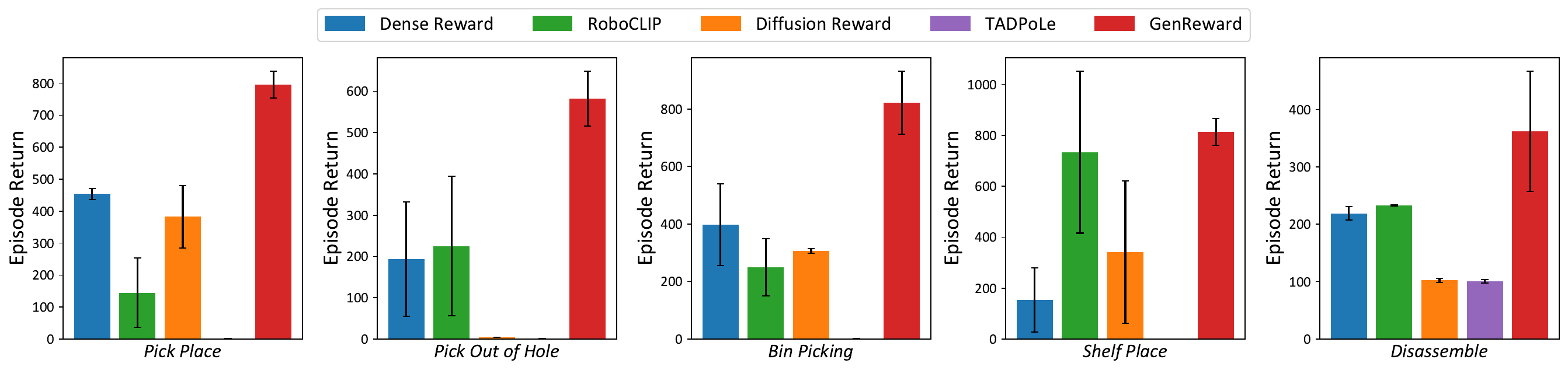}
    \vspace{-15pt}
    \caption{Performance on Meta-World complex manipulation tasks in terms of episode return under dense reward setting.}
    \label{fig:comparsion_dense}
\end{figure*}

\begin{figure*}[t]
    \centering
    \includegraphics[width=\linewidth]{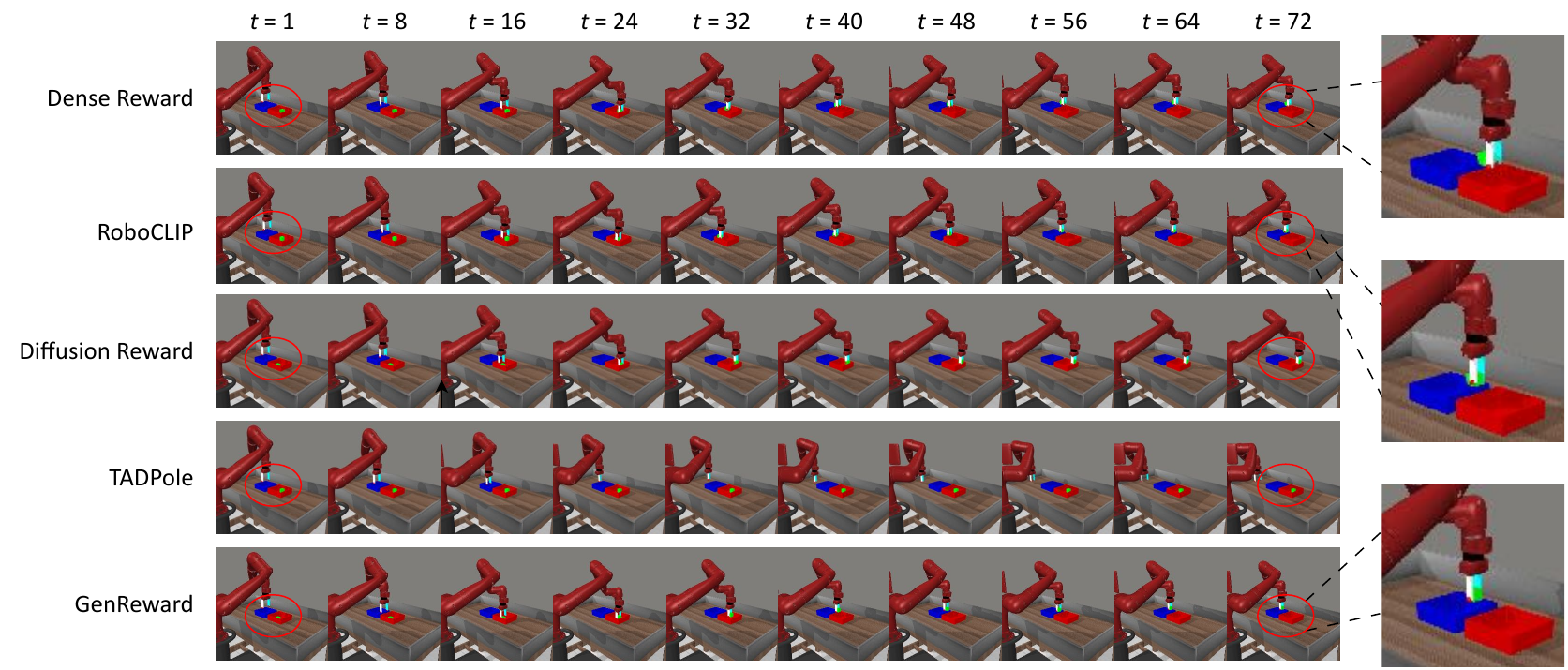}
     \vspace{-15pt}
    \caption{Policy evaluation on the Meta-World \textit{Bin Picking} task. TADPoLe fails to contact the puck, while Diffusion Reward moves the grasped puck away from the target position.
    In contrast, GenReward enables the policy to complete the grasp in fewer steps and outperforms both Dense Reward and RoboCLIP.
    }
    \label{fig:bin-pick-cmp}
\end{figure*}

\paragraph{Compared baselines.}
\label{sec:compared_baselines}
We compare \model{} with other reward models, including
% \vspace{-5pt}
\begin{itemize}[leftmargin=*]
    \item \textbf{Dense Reward}: The original dense reward provided by the environment.
    \item \textbf{RoboCLIP}~\citep{sontakke2023roboclip} leverages pretrained VLMs to calculate the similarity between the task description or demo video and historic frames as a reward for RL agents.
    \item \textbf{Diffusion Reward}~\citep{huang2024diffusion} exploits video diffusion models that were trained with the expert videos to estimate the history-conditioned entropy and utilizes its negative as rewards.
    \item \textbf{TADPoLe}~\citep{luo2024text} utilizes a pretrained text-conditioned image diffusion model to compute zero-shot dense reward, encouraging the alignment of the visual observation towards the provided text through the denoising gradient.
\end{itemize}
The performance of each reward model is evaluated by training it on top of the strong model-based RL algorithm DreamerV3~\citep{hafner2023dreamerv3}.

\subsection{Main Comparison} 

We evaluate the task performance in terms of episode return. 
\figref{fig:comparsion_dense} shows the performance of \model{} and all the baselines. 
For the Meta-World robotic manipulation tasks, we use the Bridge dataset with real-world videos as the source domain. 
We report the mean results and standard deviations over 10 episodes. As shown in \figref{fig:comparsion_dense}, our approach achieves competitive performance in episodic returns over all five tasks on Meta-World. 
Specifically, \model{} outperforms DreamerV3 with raw dense reward by a large margin in \ti{Pick Out of Hole} (193 $\rightarrow$ 582), \ti{Bin Picking} (398 $\rightarrow$ 822), and \ti{Shelf Place} (154 $\rightarrow$  814). 

\begin{figure*}[t]
    \centering
    \includegraphics[width=\linewidth]{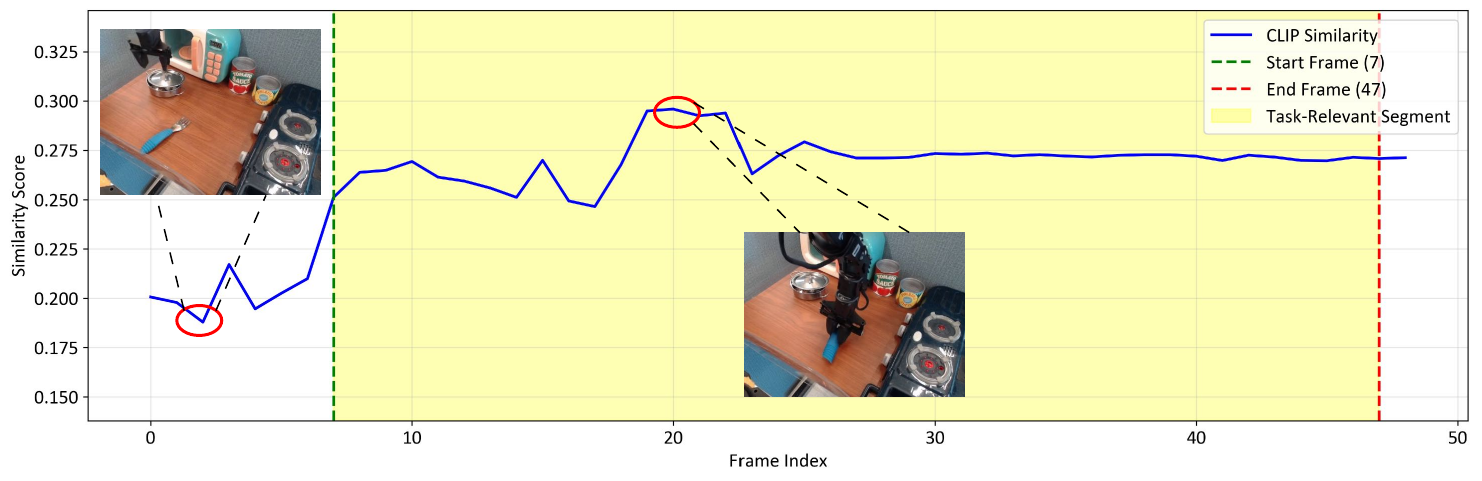}
    \caption{Showcase of selecting the goal image from the video generated with the prompt \ti{pick up the blue fork} using CLIP. The highlighted area represents the video frames that are more relevant to the task. The frame with the highest similarity reflects the frame-level goal.
    }
    \label{fig:clip_selection_res}
\end{figure*}

\begin{figure*}[t]
    \centering
    \includegraphics[width=\linewidth]{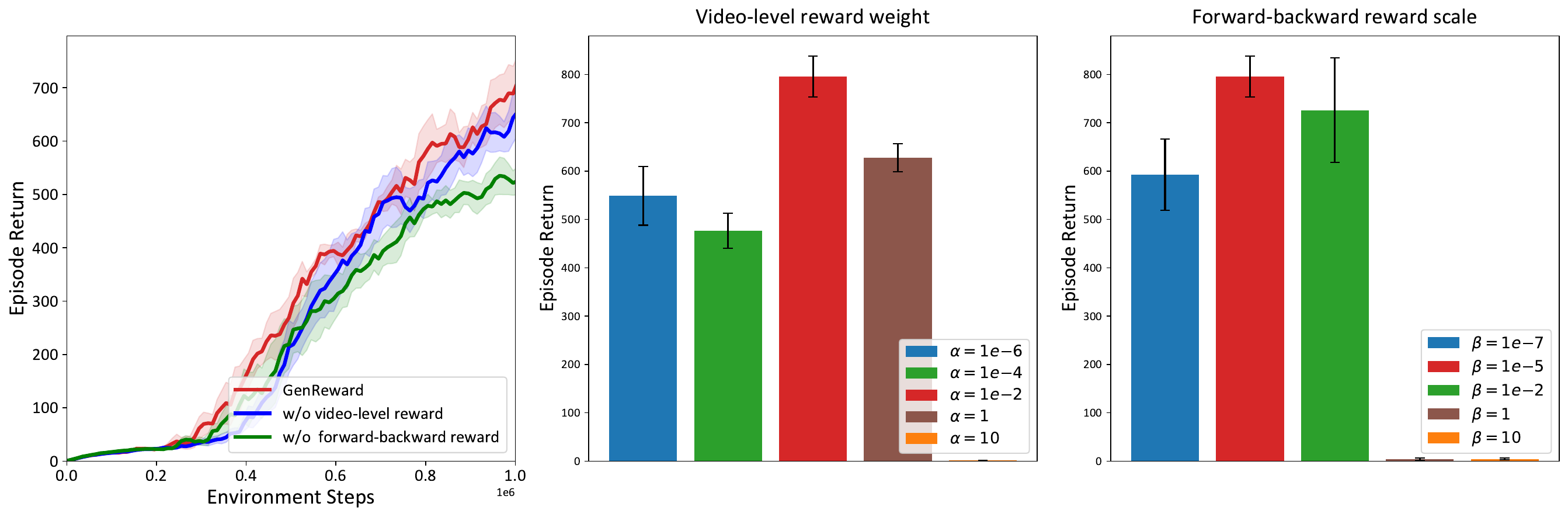}
    \caption{These figures display the ablation studies and sensitivity analyses of \model{} on Meta-World \textit{Pick Place}.
    \bb{Left:} Comparison with \model{} without video-level reward or FB reward.
    \bb{Middle:} The sensitivity analyses of video-level reward weight. 
    \bb{Right:} The performance of \model{} with different FB reward scale. 
    }
    \label{fig:ablation_sensitivity_results}
\end{figure*}
Compared to RoboCLIP and Diffusion Reward, which also employ video or diffusion-based reward, \model{} demonstrates a significant advantage by effectively exploiting and transferring the underlying world knowledge behind these videos. 
Notably, TADPoLe underperforms the other baselines in those tasks, where TADPoLe fails to provide effective rewards.
Additionally, \figref{fig:bin-pick-cmp} presents a qualitative comparison of different methods on the \ti{Bin Picking} task.
Further results on the challenging sparse reward setting are given in \appref{sec:sparse_reward}.

As depicted in \figref{fig:clip_selection_res}, we select the goal frame from the generated video based on CLIP similarity on Meta-World \ti{Pick Place}, providing a more accurate goal image for \FB{} representation learning. 

\subsection{Model Analyses} 

\paragraph{Ablation studies.}

\begin{figure}[thb]
    \centering
    \includegraphics[width=\linewidth]{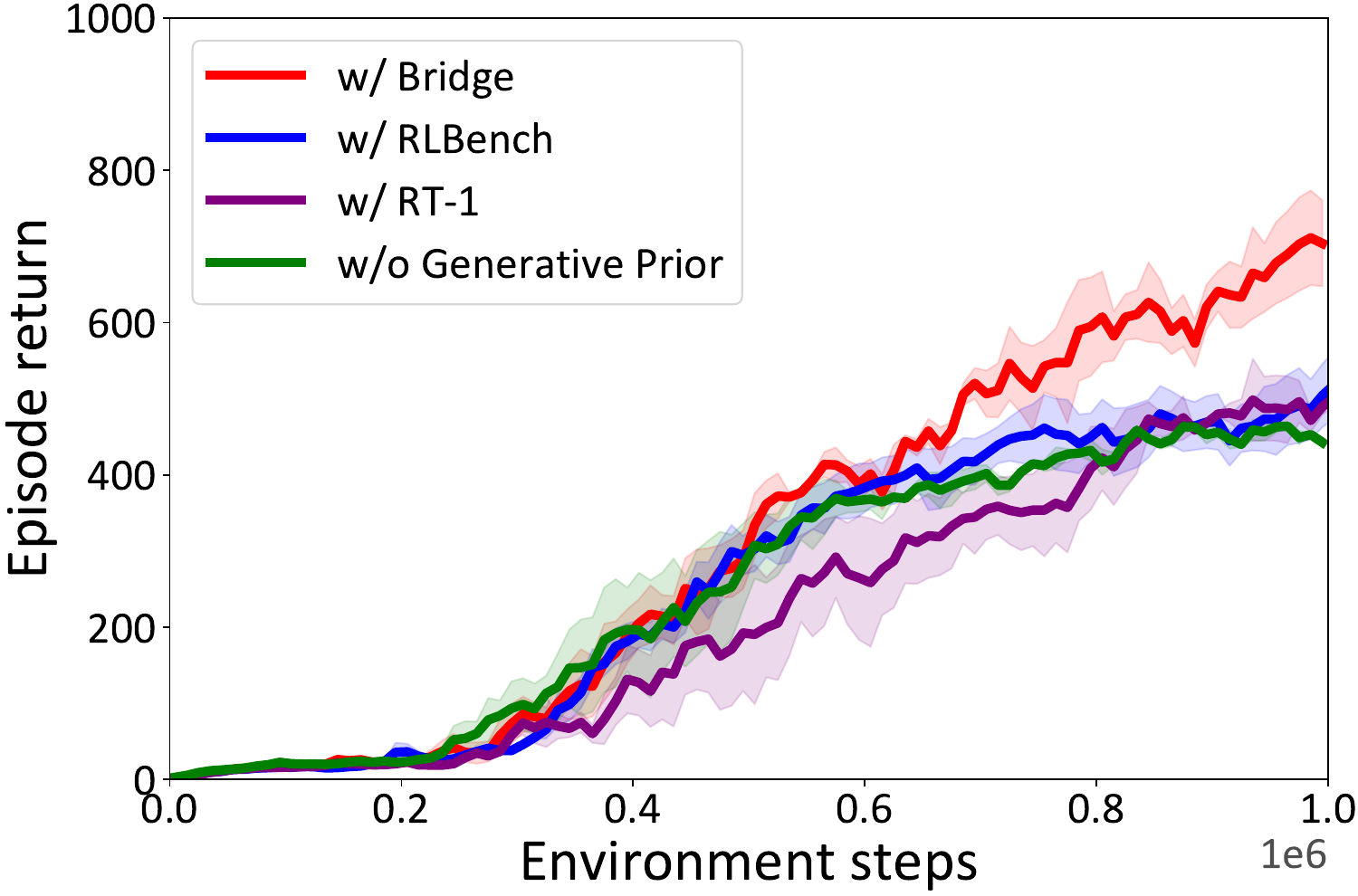}
    \caption{Performance of \model{} on Meta-World \ti{Pick Place}
    with different generated videos.}
    \label{fig:domain_selection}
\end{figure}

We conduct ablation studies to validate the effect of the video-level reward and forward-backward reward.
\figref{fig:ablation_sensitivity_results}~(Left) shows corresponding results in the $\ti{Bridge Pick Video} \rightarrow$ \textit{Pick Place}.  
As indicated by the blue curve, removing the video-level reward of \model{} results in a performance degradation, highlighting the necessity of the video-level goal. 
The green curve demonstrates that \model{} introduces frame-level goal as reward leads to superior performance.

\paragraph{Sensitivity analyses.} 
We conduct sensitivity analyses on Meta-World~(\textit{Bridge Pick Video} $\rightarrow$ \textit{Pick Place}). 
 In \figref{fig:ablation_sensitivity_results}~(Middle), we observe that when the weighting coefficient~$\alpha$ of the video-level reward is too small, the agent fails to mimic the behavior of the generated video.
In contrast, when $\alpha$ is too large, it hinders behavior learning, resulting in decreased performance.
FB reward weight $\beta$ controls the frame-level goal scale. 
When $\beta$ is set too low, the agent is unable to exploit rich world knowledge from the video diffusion models.
On the other hand, an overly large $\beta$ may cause the agent to rely too heavily on the generated frame-level goals, struggling to explore.

\paragraph{Effects of generated video domain.}
To verify the effect of the generated video domain, as shown in \figref{fig:domain_selection}, we evaluate \model{} on Meta-World \ti{Pick Place} by exploiting alternative generated videos, including frames generated from RT-1 and RLBench. 
Compared with the DreamerV3 agent without leveraging generative prior, \model{} consistently benefits from goal-driven reward, which transfers the world knowledge from the video diffusion model to the downstream agent.

% \vspace{-3pt}
\section{Related Work}
\paragraph{Reinforcement learning with diffusion model.}
A number of recent works employ diffusion models to facilitate the behavior learning of RL agents. 
DIAMOND~\cite{alonso2024diffusion} employs a continuous diffusion model based on EDM~\cite{karras2022elucidating} for world modeling, which directly models environmental dynamics in pixel space instead of operating on discrete latent sequences.
% %
%
PolyGRAD~\cite{rigter2024world} trains a diffusion model to generate an entire state-reward trajectory in a single pass, while introducing a policy score to guide the generated trajectory toward the current policy output.
TADPoLe~\cite{luo2024text} utilizes a pretrained text-conditioned diffusion model to calculate rewards for facilitating policy learning, which predicts added noise and assigns high rewards when generated frames align with the text prompt.
Diffusion reward~\cite{huang2024diffusion} adopts the negative of conditional entropy on top of a conditional diffusion model pretrained on expert data.
Different from the aforementioned approaches, our approach focuses on exploiting video diffusion model prior to provide goal-driven reward.

\paragraph{VLMs as reward function.}
Several efforts exploit VLMs as reward models have gained increasing attention in recent years.
RoboCLIP~\cite{sontakke2023roboclip} provides a sparse reward at the end of an episode by calculating the similarity between the visual observations and expert videos or text descriptions.
VLM-RMs \cite{rocamonde2024vision} leverages pretrained vision-language models as reward models for RL tasks with vision, which computes the cosine similarity between the embedding extracted from visual observations using CLIP and text prompts as rewards.
FuRL~\cite{fu2024furl} presents the fuzzy VLM reward-aided RL to mitigate the issue of reward misalignment that leads to a negative effect on the learning of the agent. 
Instead of using the VLMs, we leverage the video encoder of pretrained generative model to measure the alignment between the agent's trajectories and the generated goal videos.

\paragraph{Decision making with videos.}
Many research works have been devoted to improving decision making capability of agents with videos~\cite{baker2022video,seo2023multi,black2023zero,wu2023pre,yang2024video,bharadhwaj2024track2act,wang2025disentangled,gao2025adaworld,bu2025univla,wang2025language}.
APV~\cite{seo2022reinforcement}  adopts action-free video prediction pretraining to learn useful representations of environment dynamics, which can be finetuned for action-conditioned world modeling.
IPV~\cite{wu2023pre} introduces contextualized world models that pretrained on large-scale in-the-wild videos to improve the sample efficiency of model-based RL agents on various downstream tasks.
R3M~\cite{nair2022r3m} learns universal visual representations for manipulation by training on egocentric human videos~(Ego4D), where time-contrastive learning and video-language alignment are employed to facilitate learning performance on downstream tasks.
LIV~\cite{ma2023liv} proposes a unified framework to learn goal-conditioned visual-language value function through extending value-implicit pretraining and CLIP objective, providing zero-shot dense reward for robot manipulation.
VIPER~\cite{escontrela2023video} adopts a pretrained video prediction model to provide rewards, which considers the conditional log-likelihoods for each transition as rewards. 
UniPi~\cite{du2023learning} leverages a text-to-video diffusion model to generate a video of a desired task, while independently training an inverse dynamics model to predict the action that transitions between consecutive frames, effectively learning a goal-conditioned policy from purely synthetic data.
Luo \etal ~\cite{luo2025grounding} proposes a self-supervised framework that is based on a large pretrained video model that can provide rich priors of task completion, ground the generated video into continuous actions.
In contrast, our work adopts generated videos as video-level goals to provide rewards for goal-conditioned behavior learning.

\section{Conclusions and Limitations}
\label{sec:conclusion}
In this paper, we present a reward framework dubbed \model{}, which leverages pretrained video diffusion models as goal-driven reward models in terms of video-level and frame-level goals.
By exploiting the rich world knowledge learned by generative models, we enable RL agents to receive informative reward signals without the need for explicit reward engineering. 
Extensive experiments on Meta-World manipulation tasks demonstrate the effectiveness of our approach.

One limitation of \model{} is that we require computing video-level rewards and \FB{} rewards during training process, which introduces additional computational overhead. 

\paragraph{Acknowledgements.} 
This work was supported by Grants of NSFC 62302246 \& 62250062, ZJNSFC LQ23F010008, Ningbo 2023Z237 \& 2024Z284 \& 2024Z289 \& 2023CX050011 \& 2025Z038 \& 2025Z059, the Smart Grid National Science and Technology Major Project (2024ZD0801200), the Shanghai Municipal Science and Technology Major Project (2021SHZDZX0102), and the Fundamental Research Funds for the Central Universities.
Additional support was provided High Performance Computing Center at Eastern Institute of Technology, Ningbo, and Ningbo Institute of Digital Twin. 
We thank Haoyu Zhen and Kwanyoung Park for helpful discussions.

% WARNING: do not forget to delete the supplementary pages from your submission 

\clearpage
\setcounter{page}{1}
\setcounter{section}{0}
\setcounter{table}{0}
\setcounter{figure}{0}
\maketitlesupplementary
\appendix

\renewcommand\thesection{\Alph{section}}
\renewcommand\thetable{\Alph{table}}
\renewcommand\thefigure{\Alph{figure}}

\section{Additional Results}

\subsection{Results with Sparse Rewards}
\label{sec:sparse_reward}

\begin{figure}[t]
    \centering
    \includegraphics[width=\linewidth]{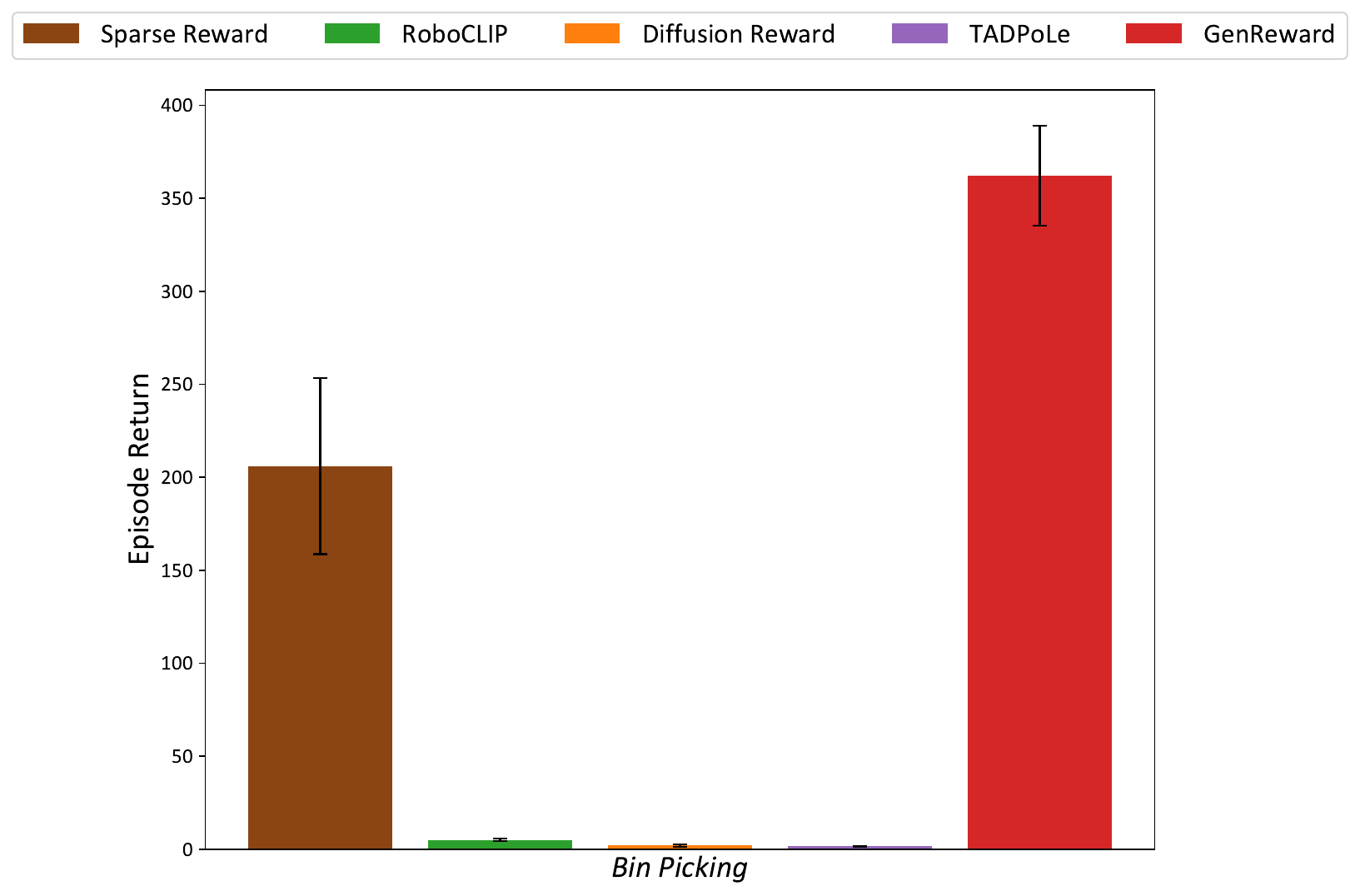}
     \vspace{-10pt}
    \caption{Performance on Meta-World \ti{Bin Picking} under sparse reward setting.
    }
    \label{fig:results_sparse}
\end{figure}

Different from dense reward settings, we consider a setup with a sparse reward function for the Meta-World tasks. 
Under this setting, the agent receives a reward every 64 steps, with the reward set to 0 before that.
We present quantitative results of sparse rewards in \figref{fig:results_sparse}.
It can be observed that \model{} still achieves consistent improvements compared to other baselines even with sparse rewards, demonstrating its effectiveness.

\subsection{Evaluation on Challenging Benchmark}
\label{sec:eval_benchmarks}
We reimplement \model{} with the model-free \ti{DrQ-v2}~\cite{yarats2022mastering} backbone on DCS and Adroit~\cite{rajeswaran2018learning} (both dense reward), and on Meta-World (limited training steps with \ti{0/1 sparse reward}), reporting \ti{success rates} for Adroit and Meta-World.
Notably, we use text-to-video generation \ti{without finetuning} on DCS.
As reported in the \tabref{tab:more_challenging_results}, \model{} consistently outperforms baselines across all benchmarks and reward settings.

\begin{table}[t]
  \caption{Performance comparison across various environments.}
  \footnotesize
  \setlength{\tabcolsep}{1.2pt}
  \centering
  \begin{center}
  \vspace{-10pt}
  \begin{tabular}{l | c c c c}
  \hline
  Model & Raw Reward & RoboCLIP & Diffusion Reward & GenReward \\
  \hline 
\ti{Walker Walk} & 640 $\pm$74  &  695$\pm$94 &  28$\pm$2 & \bb{782 $\pm$ 110} \\
\ti{Hopper Stand}   & 646 $\pm$ 161  & 589$\pm$44  &  776$\pm$58 & \bb{821$\pm$96} \\
\ti{Adroit Door} & 60$\pm$20  & 70$\pm$20 & 0$\pm$0  & \bb{90$\pm$10} \\
\ti{MW Reach }  & 25$\pm$5  & 45$\pm$5  & 60$\pm$10  & \bb{85$\pm$5} \\
% \ti{MW Drawer Open }  &  &  &  &  \\
  \hline
  \end{tabular}
  \label{tab:more_challenging_results}
  \end{center}
\end{table}

\subsection{Results on Video Hallucinations}
We do not address hallucinations. Instead, we train our method on generated videos exhibiting hallucinations (\ie, object teleportation). Interestingly, as reported in \tabref{tab:hallucination}, the agent consistently benefits from the proposed reward mechanism.

  \begin{table}[t]
  \caption{Performance with hallucination.}
  \footnotesize
  \setlength{\tabcolsep}{3.pt}
  \centering
  \begin{center}
  \vspace{-10pt}
  \begin{tabular}{l | c c c}
  \hline
  Model & Dense Reward & GenReward w/ hallucination & GenReward \\
  \hline 
  \ti{Pick Place} & 454 $\pm$30  & 574 $\pm$ 98 &  \bb{796 $\pm$ 73}\\
  \hline
  \end{tabular}
  \label{tab:hallucination}
  \end{center}
  \end{table}

\subsection{Component Sensitivity}
The results of 1) replacing CLIP with random goal image selection, 2) replacing DINOv3 with SigLIP 2 are shown in \tabref{tab:component_sensitivity}. Overall, the original design achieves the best performance among these variants

\begin{table}[t]
 \caption{Impact of individual components.}
  \footnotesize
  \setlength{\tabcolsep}{1.pt}
  \centering
  \begin{center}
  \vspace{-10pt}
  \begin{tabular}{l | c c c}
  \hline
  Model & \model{}  w/o CLIP & \model{}  w/SigLIP 2 & \model{}  \\
  \hline 
  \ti{Walker Walk} & 663$\pm$16 & 448$\pm$122 & \textbf{782 $\pm$ 110} \\
  \hline
  \end{tabular}
  % \vspace{-9.1 mm}
  \label{tab:component_sensitivity}
  \end{center}
\end{table}

\subsection{Comparison of GenReward Variants}
\tabref{tab:variant} shows that our method is competitive with the real-video variant.
Moreover, when computing the FB reward, removing the action leads to decreased performance (see \tabref{tab:variant}).
\begin{table}[t]
\caption{Performance of GenReward variants on DCS \textit{Walker Walk}.}
  \footnotesize
  \setlength{\tabcolsep}{1.pt}
  \centering
  \begin{center}
  \vspace{-10pt}
  \begin{tabular}{l | c c c}
  \hline
  Model & \model{} w/o action & \model{} w/ real & \model{} \\
  \hline 
  \ti{Walker Walk} & 435$\pm$249 & \bb{784$\pm$67} & 782 $\pm$ 110 \\
  \hline
  \end{tabular}
  % \vspace{-9.1 mm}
  \label{tab:variant}
  \end{center}
\end{table}

\subsection{VDM Adaption Details.}
Finetuning VDM takes 7 days using 16 A100 GPUs. We randomly select 400 samples from finetuning data. 
The FVD score is 21.6 indicates high fidelity of VDM.

\subsection{Computational Complexity}
 \tabref{tab:training_time} shows that, when training for 1M steps, \model{} trains faster than all baselines, except raw dense reward.
\begin{table}[t]
\caption{Training time comparison between baselines on Meta-World \textit{Pick Place}.}
   % \vspace{-15pt}
  \footnotesize
  \setlength{\tabcolsep}{0.45mm}
  \centering
  \begin{center}
  \vspace{-10pt}
  \begin{tabular}{l | c c c c }
  \hline
   Model & Dense Reward & RoboCLIP & Diffusion Reward & GenReward \\
  \hline
  Training time &  102 hours & 119 hours &  109 hours & 106 hours\\
    \hline
  \end{tabular}
  \label{tab:training_time}
  \end{center}
  \end{table}

\section{Forward-Backward Network Details }
\label{sec:fb_details}
The \FB{} objective originates from approximating the successor measure $M^{\pi_z}(s,a,s')$, which describes the discounted occupancy of future states $s'$ reachable from $(s,a)$  under the policy $\pi_z$. 
In the low-rank factorization~(see \eqref{eq:low_rank_app}), 
the inner product $F(s,a,z)^\top B(s')$ acts as a learned similarity measure: it is large if $s'$ is likely to be visited from $(s,a)$.
Minimizing the Bellman residual on this approximation, as detailed in \eqref{eq:fb_loss}, therefore encourages future states that are likely to be visited from $(s,a)$ under $\pi_z$ to receive high similarity scores in the latent space.
Additionally, the orthonormalization loss $\mathcal{L}_{\text{norm}}$ regularizes the backward representations to prevent degenerate collapse and ensure feature isotropy.
Concretely, $\mathcal{L}_{\text{norm}}=\Big\|
\mathbb{E}_\rho [B\,B^\top] - I_d
\Big\|_\tx{F}^2$.~Here $\|\cdot\|_{\tx{F}}$ is Frobenius norm.
Detailed derivation of FB loss~(see \eqref{eq:fb_loss}) can be found in \cite{touati2023does}.

\section{Effect of Frame-Level Goal Selection}
\begin{figure}[t]
    \centering
    \includegraphics[width=\linewidth]{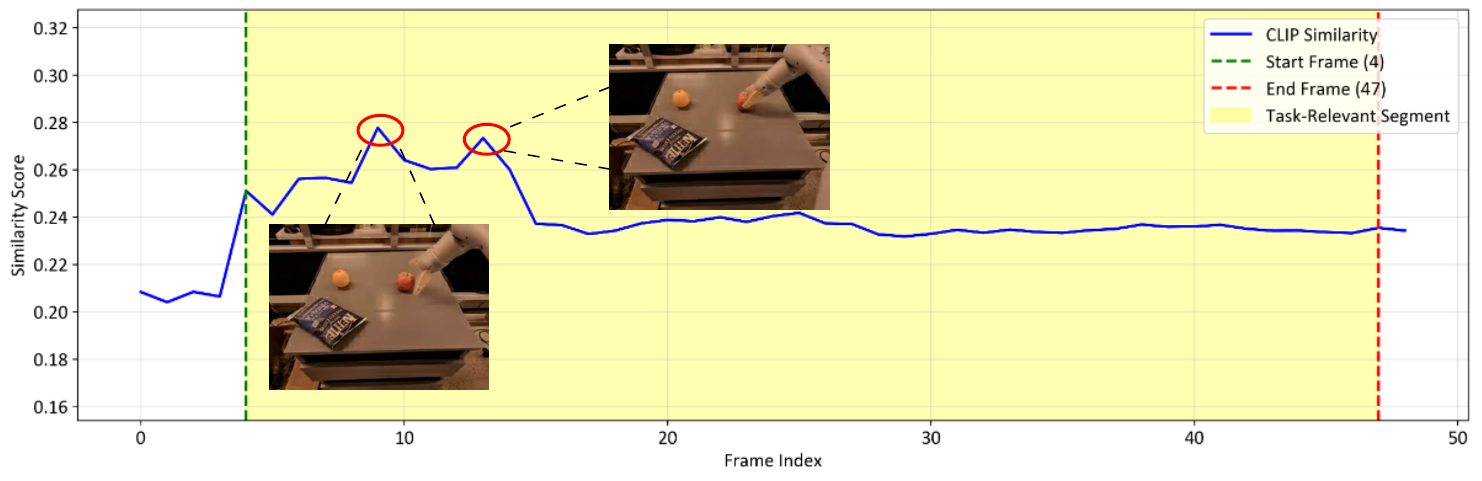}
    \caption{Failure Case of CLIP-based frame selection in a generated RT-1 \ti{Pick Apple} video. 
    The most relevant frame does not fully grasp the apple, while the second-most relevant frame actually contains a successful grasp.}
    \label{fig:clip_failure_case}
\end{figure}
As shown in \figref{fig:clip_failure_case}, in some cases, CLIP may select a frame that is not the most relevant as the goal image. 
Interestingly, \model{} with RT-1 \ti{Pick Apple} video still outperforms DreamerV3 with original reward~(see \figref{fig:domain_selection}). 
Although not the most relevant, the frame-level goal selected by CLIP can still facilitate fine-grained goal achievement of the agent.

\section{Hyperparameters}

The final hyperparameters of \model{} are listed in \tabref{tab:hparams}. 
% %
\begin{table}[H]
\centering
\caption{Hyperparameters of \model{}.} 
\vspace{-3pt}
\vskip 0.05in
\setlength{\tabcolsep}{0.6mm}{} %
\begin{tabular}{lccc}
\toprule
\textbf{Name} & \textbf{Notation} & \textbf{Value} \\
\midrule
\texttt{Video-Level Reward} \\
\midrule
Reward weight & $\alpha$ & $1\times10^{-2}$ \\
\midrule
\texttt{Forward-Backward Reward} \\
\midrule
Reward weight & \(\beta\) & \(1\times 10^{-5}\)\\
Train steps & --- & $1 \times 10^5$ \\
Observation dimension & --- & $384$ \\
Feature dim & \(d\) & $512$ \\
Hidden dim & --- & $512$ \\
Learning rate & --- & $1 \times 10^{-4}$ \\
Target network soft-update rate & --- & $0.01$ \\
\midrule
\texttt{General} \\
\midrule
Replay capacity & --- & $1 \times 10^6$ \\
Batch size & $B$ & $16$ \\
Batch length & $T$ & $64$ \\
Train ratio & --- & $512$ \\
Intrinsic reward interval & --- & \(128\)\\
\midrule
\texttt{World Model} \\
\midrule
Deterministic latent dimensions & --- & $512$ \\
Stochastic latent dimensions & --- & $32$ \\
Discrete latent classes & --- & $32$ \\
RSSM number of units & --- & $512$ \\
World model learning rate & --- & $1 \times 10^{-4}$ \\
Reconstruction loss scale & $\beta_{\text{pred}}$ & $1$ \\
Dynamics loss scale & $\beta_{\text{dyn}}$ & $0.5$ \\
Representation loss scale & $\beta_{\text{rep}}$ & $0.1$ \\
\midrule
\texttt{Behavior Learning} \\
\midrule
Imagination horizon & $H$ & $15$ \\
Discount & $\gamma$ & $0.997$ \\
$\lambda$-target  & $\lambda$ & $0.95$ \\
Actor learning rate & --- & $3\times10^{-5}$ \\
Critic learning rate & --- & $3\times10^{-5}$ \\
\bottomrule
\end{tabular}
\label{tab:hparams}
\end{table}

\end{document}